# Fitness Dependent Optimizer for IoT Healthcare using Adapted Parameters: A Case Study Implementation

Aso M. Aladdin[1,2] [0000-0002-8734-0811]
Jaza M. Abdullah[3] [0000-0002-0122-8432]
Kazhan Othman Mohammed Salih[4] [0000-0003-4394-8953]
Tarik A. Rashid[5] [0000-0002-8661-258X]
Rafid Sagban [6,7] [0000-0002-6206-7310]
Abeer Alsaddon[8] [0000-0002-2309-3540]
Nebojsa Bacanin[9] [0000-0002-2062-924X]
Amit Chhabra[10] [0000-0003-2056-6231]
S.Vimal[11] [0000-0002-1467-1206]
Indradip Banerjee[12] [0000-0002-7879-5339]

[1]Department of Applied Computer, College of Medical and Applied Sciences, Charmo University, Sulaymaniyah, KRG, Iraq. aso.aladdin@charmouniversity.org
[2]Department of Information Systems Engineering, Erbil Technical Engineering Colleg, Erbil Polytechnic University, Erbil, KRG, Iraq. aso.dei20@epu.edu.iq
[3]Information Technology, College of Commerce, University of Sulaimani, Sulaymaniyah, KRG, Iraq. jaza.abdullah@univsul.edu.iq
[4]Project Management Department, College of Commerce, University of Sulaimani, Sulaimani, 46001 KRG, Iraq. kazhan.mohammed@univsul.edu.iq
[5]Computer Science and Engineering Department, University of Kurdistan Hewler, Erbil, KRG, Iraq.
tarik.ahmed@ukh.edu.krd
[6]Department of Computer Technology Engineering, Technical Engineering College, Al-Ayen University, 64001 Thi- Qar, Iraq. rsagban@alayen.edu.iq
[7]Information Technology College, University of Babylon, Hilla, Iraq.
[8] Information Technology Department, Asia Pacific International College (APIC), Sydney, Australia. alsadoon.abeer@gmail.com
[9]Singidunum University, Danijelova 32, Belgrade, 11000, Serbia. nbacanin@singidunum.ac.rs
[10] Department of Computer Engineering and Technology, Guru Nanak Dev University, Amritsar-INDIA amit.cse@gndu.ac.in
[11]Dept of CSE Ramco Institute of Technology, Rajapalayam, Tamilnadu.
vimal@ritrjpm.ac.in
[12]Department of Computer Science, University Institute of Technology, The University of Burdwan, Burdwan, India

**Abstract**
This discusses a case study on Fitness Dependent Optimizer or so-called FDO and adapting its parameters to the Internet of Things (IoT) healthcare. The reproductive way is sparked by the bee swarm and the collaborative decision-making of FDO. As opposed to the honey bee or artificial bee colony algorithms, this algorithm has no connection to them. In FDO, the search agent's position is updated using speed or velocity, but it's done differently. It creates weights based on the fitness function value of the problem, which assists lead the agents through the exploration and exploitation processes. Other algorithms are evaluated and compared to FDO as Genetic Algorithm (GA) and Particle Swarm Optimization (PSO) in the original work. The key current algorithms—The Salp-Swarm Algorithms (SSA), Dragonfly Algorithm (DA), and Whale Optimization Algorithm (WOA) have been evaluated against FDO in terms of their results. Using these FDO experimental findings, we may conclude that FDO outperforms the other techniques stated. There are two primary goals for this chapter: first, the implementation of FDO will be shown step-by-step so that readers can better comprehend the algorithm method and apply FDO to solve real-world applications quickly. The second issue deals with how to tweak the FDO settings to make the meta-heuristic evolutionary algorithm better in the IoT health service system at evaluating big quantities of information. Ultimately, the target of this chapter's enhancement is to adapt the IoT healthcare framework based on FDO to spawn effective IoT healthcare applications for reasoning out real-world optimization, aggregation, prediction, segmentation, and other technological problems.

## 4.1 Introduction

FDO is nothing more than an attempt to emulate the behavior of bees during reproduction. As a scout bee, this algorithm is designed to simulate scout bees using this strategy to choose a new home among the many colonies that are around. A proposed solution to this algorithm is a scout bee that looks for fresh hives; also, selecting the best hive among a large number of good hives is regarded to be approaching optimality [1].

Swarming is an early occurrence that develops when a fresh colony of honeybees is generated. The scout bees leave the previous hive, and the queen bee stays with a group of honeybees; Figure 4.1 illustrates the cycle of bee swarming. A swarm is made up of hundreds to tens of thousands of bees. The scout bees will briefly dwell twenty to forty meters outside from the birth beehive for several hours to a few days or weeks.

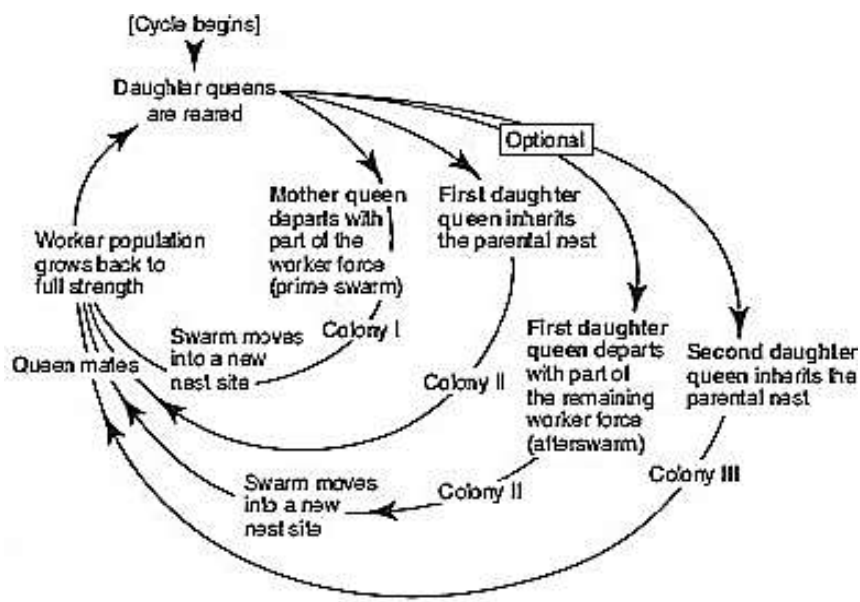

Figure 4.1. Bee swarming process cycle [2].

The colony might contribute attributes to improve this operation. A controlled honeybee colony's health has been reflected in colony characteristics. External factors influencing health and colony production express the production of an organized beehive colony. Bee brand construction and fertilization services are added since they are the driving factors for beekeepers' decision to preserve a honeybee colony [3]. Figure 4.2 includes the diagram to demonstrate the definitions of the elements as well as the interactions between them.

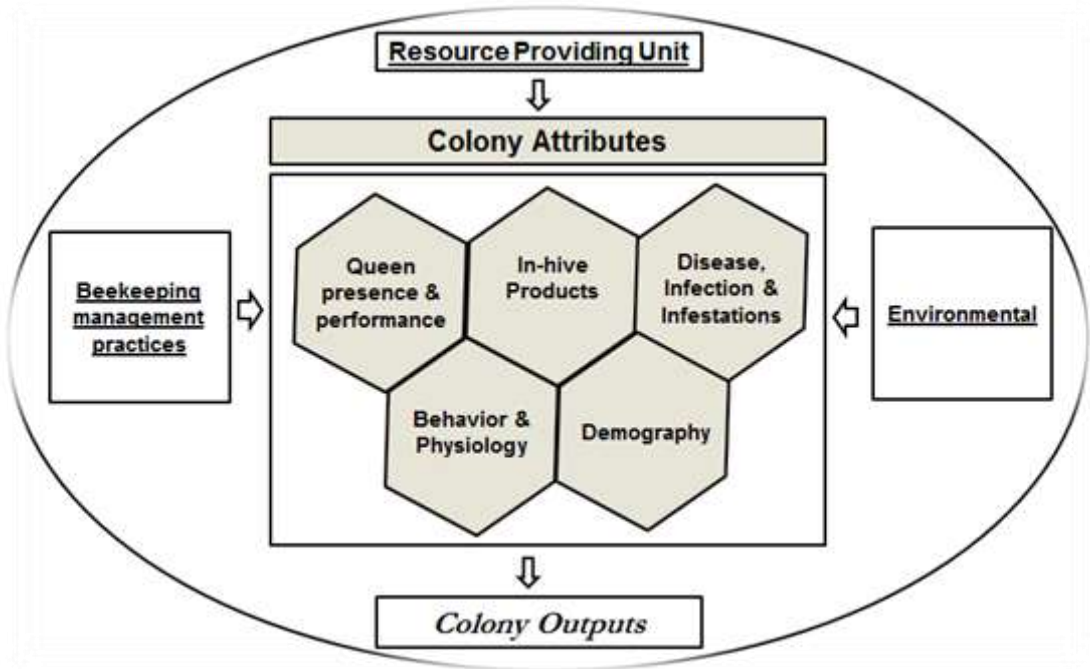

Figure 4.2. A comprehensive evaluation of the health of a honeybee colony management will take into account colony attributes (includes filling in gray), outside drivers (underlined parts in squares), and colony products (Italic written in a rectangle) [3].

IoT is an area in which massive amounts of data are transferred continuously, and FDO, yet still very new, is very powerful and provides promising results, therefore, the authors of this chapter, believe that this algorithm needs step by step explanation so that the reader can have a better understanding, the above can be the main motive behind writing this chapter. Consequently, this chapter will also introduce the improved FDO for IoT Healthcare using adapted parameters; it is a real case study implementation, which is related to the potential IoT applications. The primary goal of this chapter is to provide a step-by-step implementation guide for the FDO and provide IoT healthcare [4] as a probable application with adapting FDO parameters.

The shortened major contributions of this chapter are presented two crucial points; it is illustrated the steps calculation of the novelist swarm intellectual algorithm; besides, it detailed how to use an optimization technique to provide appropriate weights to support the methods in both the explorative and exploitative stages. The second point is concentrated on adapting the parameters of FDO for using IoT Healthcare to invent the best accumulation, expectation, and segmentation in the innovative model IoT applications. Some other unique aspects of FDO that might be improved for this reason is that it keeps earlier search agent pace for probable reprocess in consequent stages.

The following is how this chapter is organized: The second section is devoted to identifying IoT healthcare technologies. The third section is stated on IoT applications in metaheuristic algorithms and improved FDO parameters in the IoT healthcare system. The fourth section examines the mathematical formula, which can be explained for the FDO algorithm, and demonstrates the formulations step by step. The fifth section is devoted to explaining FDO by simple example as a case study. The chapter also includes a conclusion and proposes the significant feature works.

## 4.2 IoT Healthcare Technology

IoT is gradually making its way into a wide range of businesses, from manufacturing to healthcare, communications, and even agriculture [5, 6]. The IoT has looked promising in integrating a range of sensors, medical equipment, and healthcare professionals to carry high-value health treatment facilities to those in isolated locations. Using cloud-based IoT in healthcare allows for a smart healthcare system because of its large storage capacity [7]. It has enhanced patient safety, minimized healthcare costs, expanded healthcare service accessibility, and raised operational efficiency in the healthcare business [8]. The technical advancements accumulated over the past have now enabled the detection of various diseases and monitoring systems utilizing tiny technologies, such as smartwatches. Likewise, technological advancements have converted a clinic healthcare system into a physician system [9]. Sensors and IoT-enabled hospital instruments securely transmit sensitive healthcare data to healthcare specialists who might just review and take relevant steps if required without the need for human participation [10]. The technologies employed to construct an IoT of Healthcare system are important for this goal. This is because incorporating certain technologies into an IoT system might enhance its capabilities. Hence, a variety of tool technologies have been used to integrate multiple healthcare applications with an IoT structure [11]. IoT is transforming the healthcare industry by changing the way devices and people interact in healthcare service provision. The IoT implementation in healthcare will make life easier for everyone involved, including doctors and patients, while also enhancing the quality of care [12,13,14]. IoT sensors collect patient data, which is then analyzed using machine learning techniques [15]. For patients, it is simple to access and use sensors that are built into their health monitoring devices [16,17]. A doctor, for example, can monitor a patient's heart rate from their office using sensors that periodically collect data [16,18]; thus, doctors can keep tabs on and communicate with their patients while they are away from the hospital.

Monitoring patients in remote areas, telemedicine, and health technologies are just a few examples of how the IoT might very well support healthcare [19]. In healthcare, IoT has applications that are also beneficial for physicians, patients, hospitals, families, and insurance companies [20]. Accordingly, IoT healthcare technologies are categorized into three parts: identity technology, communication technology, and location technology [21]. The massive amounts of data produced by these associated devices can revolutionize healthcare. Analog data is typically acquired through sensors and other equipment. Next, the data are standardized and pre-processed before they're shipped off to a data center. At the essential level, final data is controlled and evaluated [22]. IoT applications for advanced analytics are revolutionizing healthcare by assuring better care, improving treatment effects, and lowering costs for patients, as well as controlling procedures and operations, improving efficiency, and a bettering patient experience for healthcare professionalization [23]. Using IoT devices and applications in the healthcare industry is shown in Figure 4.3 [16, 24].

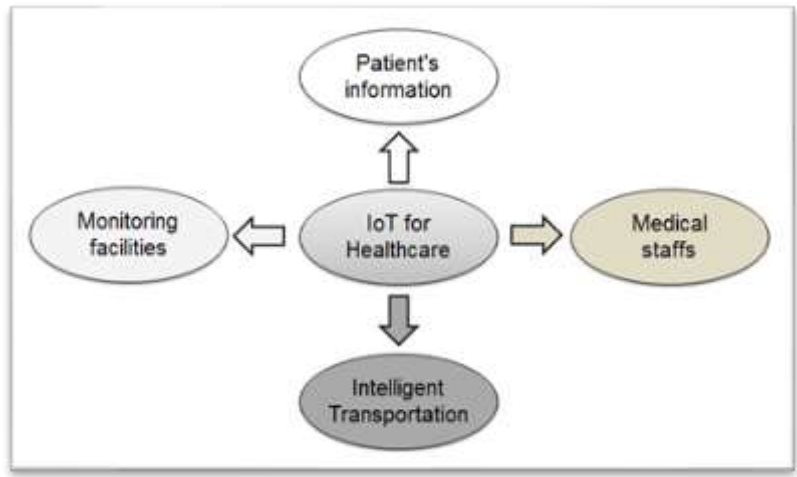

Figure 4.3. IoT for healthcare [8,15].

## 4.3 IoT Applications in Meta-heuristic Algorithms

IoT applications can be adapted in meta-heuristic algorithms to simply obtain the behaviors of big data. To make IoT networks last longer, several solutions have been developed; Quantum particle swarm optimization (QPSO) is a meta-heuristic strategy that researchers have presented earlier as a method for lengthening the lifespan of the IoT network using the Cooperative Multiple Input Multiple Output system or so-called MIMO.

At the termination of each stage, researchers aim to select dynamically the ideal cooperative MIMO transmitter and receiver equipment that causes a long operational lifecycle. Researchers have devised a strategy for finding the best cluster headers in another study [25]. Battery life, storage capacity, and communication range are all significantly reduced in sensor nodes. To transmit data to IoT devices, the author used the suggested communication protocol [26]. Whale Optimization Algorithm is employed in another meta-heuristic study to construct the ideal energy-conscious head collection for the Wireless Sensor Network–Internet of Things (WSN-IoT). The dormancy, workload, energy, length, and thermometer were all taken into description when making the final decision on the best Cluster Head feature to deploy. Algorithms; such as ABC, PSO, GA, and GSA with WOA-based Cluster Head selection methods or adaptive GSA; are compared to the proposed model [27].

The study presented in Augusto et al. investigated ways of controlling the behavior and conditions of nocturnal persons in a healthcare system [28]. It has been suggested by Tomar et al. that data mining algorithms can be used to provide a more accurate forecast, as an example, to improve the healthcare system's ability to predict human behavior. Even though typical data mining algorithms might produce significant outcomes for the healthcare system, this research via Tomaret et al., Tsai et al., and Yoo et al. has also advocated that the use of meta-heuristics can help improve the outcomes of healthcare system analytics [29, 30, 31].

Because a healthcare system's large data can comprehend a significant collection of potentially useful information, scholars have been interested in learning. Besides, in recent decades, there has been a lot of debate on how to develop a high-performance algorithm for data mining analysis or data science inquiries [32]. As a result, we recommended this parallel meta-heuristic algorithm for analyzing the IoT healthcare big data mining methods. IoT and Artificial intelligence; on the other hand; are poised to transform most industries, but possibly not any more so than healthcare. Furthermore, both biomedical and computer vision technologies may analyze data collected in public healthcare databases to detect and improve care issues [33, 34]. But from the other hand, healthcare IoT is concerned with medical equipment connectivity, where recorded data may be saved and evaluated for future diagnostic procedures employing intelligent approaches [35]. Notwithstanding this, this chapter demonstrates that FDO is a swarm technique that provides alternative ideas for analyzing health data and detects abnormalities to increase life quality.

According to the above reasoning, it is possible to infer that, even though there are various models available, they all have significant power consumption. To solve this issue, a case study of the FDO algorithm can be utilized in IoT healthcare to optimize power consumption by choosing the optimum cluster.

Thus, the FDO algorithm can provide a brief analysis of meta-heuristics for the healthcare system as well as a road map for academics working on meta-heuristics and healthcare to produce a more proficient and productive health insurance system. This technique offers a learnable data analytics framework that may be collected. Then, the data may be presented with a possible solution by FDO parameters to the incorporation of diverse forms of input data from various devices, sensors, and equipment in addition to a simple evolutionary result to the gigantic data dilemma that a healthcare system would be confronted.

## 4.4 Mathematical Formulation for FDO Algorithm

The basis of formulating the FDO problem is mainly concerned with the minimization of real searching for more positions. Scout bees, according to FDO, hunt for better hives by randomly visiting more sites, as seen in Figure 4.1; the previously discovered hive is disregarded once a superior hive is located. As a result, whenever the engine discovers a new candidate, the preceding exposed solution is rejected when FDO discovers the best new solution. Secondly, if the current motion fails to find a better solution for a main artificial scout bee (hive), it reverts to its former course in the hopes of finding a better option. If the prior path fails to yield a better result, it relapses to the earlier solution, which is the premium answer available at the time.

Likewise, this FDO framework, which can be utilized in the applications of healthcare, helps to integrate the benefits of cloud computing with IoT into the medical sector. Patients' data from sensors and medical equipment can also be transmitted using the specified methods.

IoT also randomizes and calculates the financial benefits and costs of mass implementation of the record systems of electronic health, as well as simulates crucial healthcare and safety benefits depending on the big data [36, 37]. Scout bees, therefore, scout for hives at random in the wild. When adopting this method, artificial scouts initially randomly travel over the landscape to gather information about the terrain's topography. So when an artificial scout bee speeds up its current location, it expects to find a better option.

Firstly, the procedure begins by generating a random scout colony in the search space $X_i$ ($i$ is indicated the scout bee as inhuman (artificial) and *Pace* which is the path of the scout bee and the drive rate. Depending on the fitness weight *(fw)*, (*Pace*) is indicated. Also, $i$ signifies the presence of the search agent (bee) and $t$ signifies the most latest iteration. The behavior of artificial scout bees is stated in equation (4.1):

$$X_{i,t+1} = X_{i,t} + pace \qquad (Eq. 4.1)$$

As stated, the direction of *pace* is dependent on an arbitrary mechanism. The *fw* worked on concerning with minimizing problems and it could be formulated in equation (4.2).

$$fw = \left|\frac{x^*_{i,tfitness}}{x_{i,tfitnees}}\right| \qquad \text{(Eq. 4.2)}$$

The best global solution for fitness function value is $x^*_{i,t\,fitness}$, that has been exposed from distance. The rate of the current solution for fitness function is specified by $x_{i,t\,fitness}$ and $wf$ is a weight factor that is predicted either 0 or 1 and is used for adjusting the $fw$.

Depending on formula (4.2), $fw$ result identified to 0 and 1, can be neglected because it denotes a high and low chance of convergence. Occasionally, As the fitness function cost is reliant on optimization problems, thus, the opposite case occurs. Though the range $fw$ value ought to be between [0, 1]; particular situations will be $fw = 1$, which is the best global solution. It signifies that the present and best global solutions are indistinguishable, or that they have equivalent fitness values. Furthermore, it is plausible that $fw = 0$, which happens when $x^*_{i,tfitness} = 0$. the rules can be illustrated in formulas (4.3), (4.4), and (4.5).

$$\begin{cases} fw = 1 \; or fw = 0 \; or \; x_{i,tfitness} = 0, \quad pace = x_{i,t} * r & \text{(Eq. 4.3)} \\ fw > 0 \; and fw < 1 \begin{cases} r < 0, pace = (x_{i,t} - x^*_{i,t}) * fw * -1 & \text{(Eq. 4.4)} \\ r \geq 0, \quad pace = (x_{i,t} - x^*_{i,t}) * fw & \text{(Eq. 4.5)} \end{cases} \end{cases}$$

There are other arbitrary pace implementations; however, Levy flight was elected since it gives supplementary steady motions due to its excellent distribution curve [38]. $r$ is a number randomly generated between the range.

If the preceding pace does not lead the scout bee to the best solution, the optimizer will keep the existing solution until the next round. When the answer is accepted in this procedure, the *Pace* value is kept for possible reprocess in the subsequent iteration. Two minor deviations are required for implementing this algorithm for maximization problems. Equation (4.2) obliges to be firstly substituted by equation (4.6), which is simply the opposite of equation (4.2).

$$fw = \left|\frac{x_{i,tfitness}}{x^*_{i,tfitness}}\right| - wf \qquad \text{(Eq. 4.6)}$$

The criteria for picking an improved result should be secondly modified. The condition "if ($X_{t+1,i}$ fitness $< X_{t,i}$ fitness)" required substituting with the condition "if ($X_{t+1,i}$ fitness $> X_{t,i}$ fitness)". In Figure 4.4, The SOFDO pseudo-code has displayed both rates.

```
Initialize scout bee population X_{t,i} (i = 1, 2, ..., n)
while iteration (t) limit not reached
    for each artificial scout bee X_{t,i}
        find best artificial scout bee x*_{t,i}
        generate random walk r in [-1, 1] range
        if( X_{t,i} fitness == 0) (avoid divide by zero)
            fitness weight = 0
        else
            calculate fitness weight. equation (4.2)
        end if
        if (fitness weight = 1 or fitness weight = 0)
            calculate pace using equation (4.3)
        else
            if (random number >= 0)
                calculate pace using equation (4.5)
            else
                calculate pace using equation (4.4)
            end if
        end if
        calculate X_{t+1,i}    equation (4.1)
        if( X_{t+1,i} fitness < X_{t,i} fitness)
            move accepted and pace saved
        else
            calculate X_{t+1,i} equation (4.1)   with previous pace
            if (X_{t+1,i} fitness < X_{t,i} fitness)
                move accepted and pace saved
            else
                maintain current position (don't move)
            end if
        end if
    end for
end while
```

Figure 4.4  Pseudocode of SOFDO [1]

## 4.5 Case Study Implementation

Researchers should have acquired numerous lessons about how to formulate models effectively and what sort of algorithm would solve these issues efficiently and reliably after researching various linear or nonlinear optimization problems [39]. While a single example cannot be used to deduce these teachings, it may be used to illustrate them. The train problem for the FDO method is described and summarized for researchers to state succinctly and applied for different aspect models in the future, particularly in the realm of IoT healthcare.

As discussed in the previous section, use the FDO algorithm for minimizing real searching positions. Table 4.1 is the random problem, which includes two dimensions and the upper bound ended at (100) with the lower bound limited at (-100).

Consequently, the example problem dimensions equal TWO, which means *x1* and *x2* for every Bee in FDO. It can be suggested three dimensions, which means *x1*, *x2*, and *x3*. This example uses two features as a bee pace for IoT healthcare.  In several healthcare applications, such as optical, electrochemical, or material physical properties, it might be used as an active sensing element or as a providing substrate [40].

Table 4.1. Fifteen random sequences for two parameters

| No. of sequences | parameter 1 | Parameter2 | No. of sequences | parameter 1 | Parameter2 | No. of sequences | parameter 1 | Parameter2 | No. of sequences | parameter 1 | Parameter2 | No. of sequences | parameter 1 | Parameter2 |
|---|---|---|---|---|---|---|---|---|---|---|---|---|---|---|
| 1 | -0.49 | -0.47 | 11 | -0.09 | 0.92 | 21 | 0.34 | -0.48 | 31 | 0.64 | 0.41 | 41 | 0.77 | -0.93 |
| 2 | 0.44 | -0.78 | 12 | 0.6 | -0.45 | 22 | -0.72 | 0.84 | 32 | -0.19 | -0.49 | 42 | 0.48 | 0.31 |
| 3 | -0.22 | -0.52 | 13 | -0.15 | 0.4 | 23 | 0.82 | -0.15 | 33 | 0.48 | 0.31 | 43 | -0.19 | 0.53 |
| 4 | 0.82 | -0.22 | 14 | 0.36 | -0.47 | 24 | -0.4 | 0.67 | 34 | -0.04 | -0.83 | 44 | 0.58 | 0.71 |
| 5 | -0.81 | 0.82 | 15 | -0.25 | 0.61 | 25 | 0.77 | -0.93 | 35 | 0.58 | 0.31 | 45 | -0.18 | 0.27 |
| 6 | -0.94 | -0.29 | 16 | 0.17 | -0.55 | 26 | -0.97 | 0.62 | 36 | -0.1 | 0.27 | 46 | 0.66 | -0.26 |
| 7 | -0.10 | 0.95 | 17 | -0.88 | -0.64 | 27 | 0.8 | -0.84 | 37 | 0.15 | -0.26 | 47 | 0.58 | 0.64 |
| 8 | 0.12 | -0.96 | 18 | 0.13 | 0.94 | 28 | -0.28 | 0.63 | 38 | -0.62 | 0.38 | 48 | -0.12 | 0.37 |
| 9 | -0.89 | 0.08 | 19 | -0.25 | -0.35 | 29 | 0.38 | -0.55 | 39 | 0.42 | -0.4 | 49 | 0.15 | -0.26 |
| 10 | 0.68 | -0.06 | 20 | 0.08 | 0.63 | 30 | -0.94 | -0.68 | 40 | 0.72 | 0.52 | 50 | -0.98 | -0.65 |

The study selected three bees as suggested by the inquiry, which means the population size equals three. Conferring to FDO, finding fitness function needs the formula to point out the pace between populations. $F_{(X)} = \sum_{i=1}^{D} X_i^2$ is generated to calculate the fitness function for this case study.

$$Bee\text{-}x1 = r * Upper\text{-}bound$$
$$Bee\text{-}x2 = r * Lower\text{-}bound$$

FDO needs a list of stochastic numbers to evaluate the behaviors. In Table 4.1, use this list of random numbers for generating random numbers respectively, and use each cell only once. The sample of study reaches the algorithm working for two iterations.

### 4.5.1 Calculating First Iteration

Regarding FDO, Bee one is represented as *B1* and the following are the initial stats that may be found:
  *B1_x1 = -0.49 \* Upper-bound = -0.49 \*100 = -49*
  *B1_x2 = -0.47 \* Lower-bound = -0.47 \* -100 = 47*

Formerly, it must be finding fitness function optimization according to the given equation $F_{(X)} = \sum_{i=1}^{D} X_i^2$, and it could be obtained this fitness

$$x_1^2 + x_2^2 = (-49)^2 + (47)^2 = \mathbf{4610}$$

Then, Bee two is symbolized as B2 and could be found initial stat as follow:

> B2_x1 = 0.44 * Upper-bound = 0.44 *100 = 44
> B2_x2 = -0.78 * Lower-bound = -0.78 * -100 = 78

Consequently, it must be also found fitness function optimization according to the same given equation $F_{(X)}= \sum_{i=1}^{D} X_i^2$, and it could be obtained this fitness for the second bee, as follow

$$x_1^2 + x_2^2 = (44)^2 + (-78)^2 = \mathbf{8020}$$

Finally, Bee three is symbolized as *B3* and could be found initial stat the same as *B1* and *B2*. By way of:

> B3_X1 = -0.22 * Upper-bound = -0.22 *100 = -22
> B3_X2 = -0.52 * Lower-bound = -0.52 * -100 = 52

It must be also finding fitness function optimization according to the same given equation $F_{(X)}= \sum_{i=1}^{D} X_i^2$, and this fitness for the third bee is given as follows:

$$x_1^2 + x_2^2 = (-22)^2 + (52)^2 = \mathbf{3188}$$

As a result, the solution of the third bee is defined as the global best solution because it has the lowest fitness value according to the FDO standard and then jumped for the second iteration. To illustrate the calculation results in the first iteration steps following up on Table 4.2 and Figure 4.5.

Global best Solution for the first iteration: $x^*_{i,t\ fitness} = \mathbf{3188}$    [-11, 52]

Table 4.2.  the first iteration steps to Find Global Solution

| Steps | Pace 1 | Pace 2 | Fitness Function Solution |
|---|---|---|---|
| 1 | - 49 | 47 | 4610 |
| 2 | 44 | -78 | 8020 |
| 3 | - 22 | 52 | 3188 |

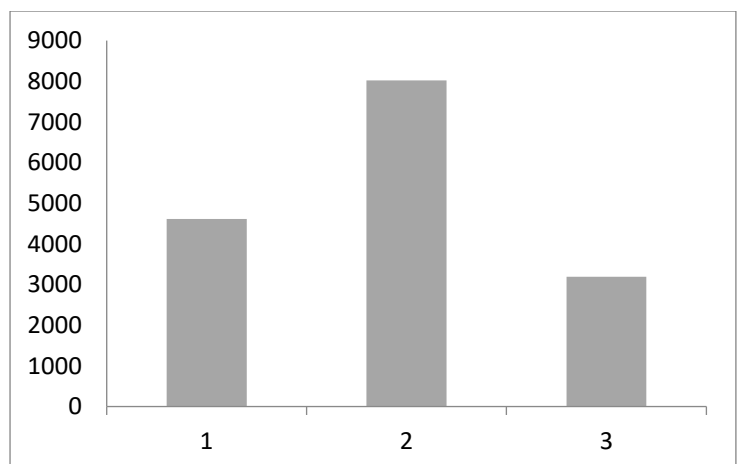

Figure 4.5.  prove Step three is the best global solution for the first iteration

## 4.5.2 Calculating Second Iteration

In the second iteration, it should first find the behavior of artificial scout bees by the following clearance steps, however, respecting the rules in the equation (4.3), (4.4), and (4.5) to find fitness weight, as shown at Figure 4.6.

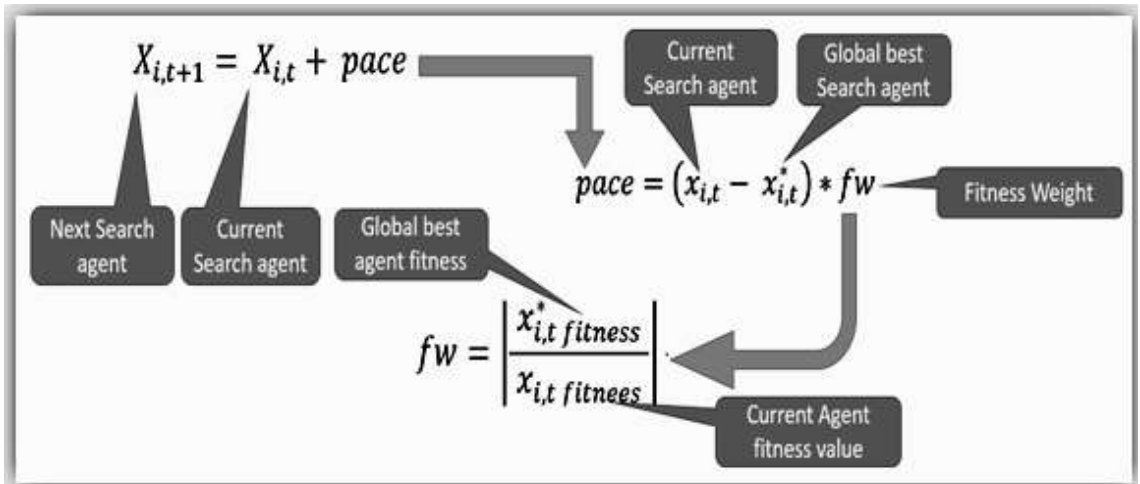

Figure 4.6. Implementing FDO sequence rules

It could be found fitness weight as follow to calculate *Pace*:

$$fw = \left|\frac{x^*_{i,tfitness}}{x_{i,tfitnees}}\right| = \frac{3188}{4610} = 0.69$$

After that the pace must be calculated according to the $r$ is a number generated randomly between the range [1-, 1] as stated in previous and modifying the scout bee location as the simple calculation:

When $r = 0.82$ then $pace = (x_{i,t} - x^*_{i,t}) * fw$,
So, B1_x1_pace = ((-49) – (-22)) * 0.69 = -33.82
When $r = -0.22$ then $pace = (x_{i,t} - x^*_{i,t}) * fw * -1$,
So, B1_x2_pace = (47 - 52) * 0.69 * -1 = 3.45

Then similar previous iteration; could be found a fresh state for Bee one; the formula (4.3) is needed as follow:

$$B1\_x1 = x_{1,1} + pace = -49 + (-33.82) = -82.88$$

$$B1\_x2 = x_{1,2} + pace = 47 + (3.45) = 50.45$$

The same given equation used in the first iteration is formulated for the second iteration. Thus, the result shows that new fitness is smaller than current fitness; as a result, it could be used for getting new weight for the second bee.

Fitness value $= x_{i,t\ fitnees} = x_1^2 + x_2^2 = (-82.88)^2 + (50.45)^2 =$ **9414**

Since, $x_{i,t\ fitnees} = \mathbf{9414} > x^*_{i,t\ fitness} = 3188$

As a result, Define Global's best Solution: $x^*_{i,t\ fitness} = \mathbf{3188}$    **[-22, 52]**

For continuity of this iteration and to calculate the fitness function for the second Bee, it should find fitness weight as follow to calculate Pace for the second Bee:

$$fw = \left|\frac{x^*_{i,t\ fitness}}{x_{i,t\ fitness}}\right| = \frac{3188}{8020} = 0.397$$

When $r = -0.81$ then $pace = (x_{i,t} - x^*_{i,t}) * fw * -1$,

So; *B2_x1_pace = (44 – (-22))* 0.397* -1= - 26.20*

When $r = 0.82$ then $pace = (x_{i,t} - x^*_{i,t}) * fw$,

So; *B2_x2_pace = (78 – 52)* 0.397 = 10.32*

Also, depending on the formula (4.3) is needed; we will calculate a new state for Bee two as follow:

$$B2\_x1 = x_{2,1} + pace = 44 + (-26.20) = 17.8$$
$$B2\_x2 = x_{2,2} + pace = 78 + 10.32 = 88.32$$

Compute the value of fitness via the same fitness function and evaluate it with the current value, the result shows as follows:

Fitness value = $x_{i,t\ fitness} = x_1^2 + x_2^2 = (17.8)^2 + (88.32)^2 = \mathbf{8117.26}$

Since, $x_{i,t\ fitnees} = 8117 > x^*_{i,t\ fitness} = 3188$

Hence, the best global value must be defined as the previous step and remain the identical fitness solution: $x^*_{i,t\ fitness} = \mathbf{3188}$    **[-22, 52]**

In the termination, this iteration ended by computing the fitness function for Third Bee. The first step for Bee three ought to catch fitness weight:

$$fw = \left|\frac{x^*_{i,t\ fitness}}{x_{i,t\ fitnees}}\right| = \frac{3188}{3188} = 1$$

When $r = -0.94$ then $pace = (x_{i,t} - x^*_{i,t}) * fw * -1$;

So; B3_x1_pace = ((-22) – (-22))*(1 * -1) = 0

When $r = -0.29$ then $pace = (x_{i,t} - x^*_{i,t}) * fw * -1$;

So; B3_x2_pace = (52 – 52)* (1 * -1) = 0

Then finding the fitness value and compare with the current value, the result shows as follows:

$$\text{Fitness value} = x_{i,t\ fitnees} = x_1^2 + x_2^2 = (0)^2 + (0)^2 = 0$$

$$x_{i,tfitnees} = 0 < x^*_{i,tfitness} = 3188$$

Here, it should be selected and defined as the global best solution. According to the FDO rules, the novel fitness solution is progressed and should be used in the next iteration. So, the new fitness Global best Solution is: $x^*_{i,t\ fitness} = 0\ [0, 0]$, besides the steps are explained observably in Table 4.3 and more illustrated in Figure 4.7.

Table 4.3. Second iteration steps to Find Global Solution

| *Steps* | *Pace* 1 | *Pace* 2 | Fitness Function Solution |
|---|---|---|---|
| **1** | -82.88 | 50.45 | **9414** |
| **2** | 17.8 | 88.32 | **8117** |
| **3** | 0 | 0 | **0** |

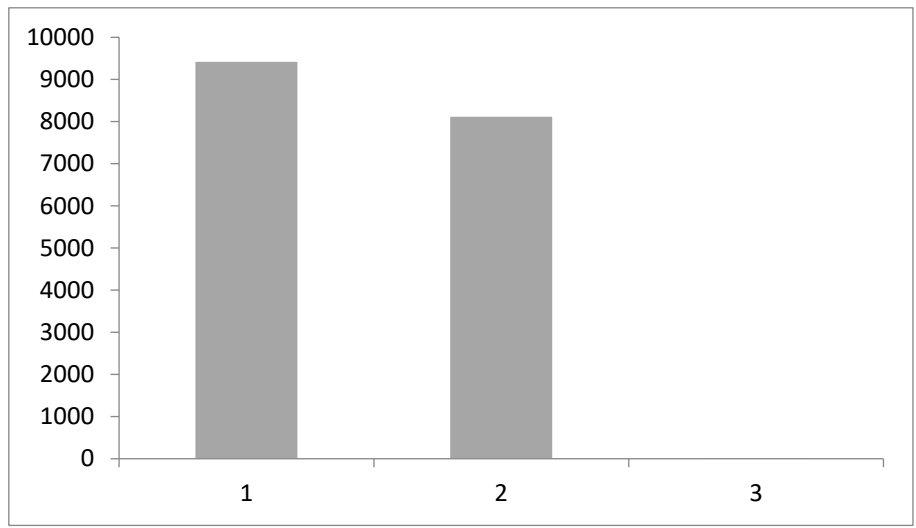

Figure 4.7. prove step three is the best global solution for the second iteration

The case study will be continued for the next iterations to pick up the best global solution according to FDO. Thus, the actual problem is about some iterative algorithm. The researcher wants to invent the appropriate parameter (s.t.). As a result; the algorithm terminates in minimal iterations. These parameters, which include scout bee in the FDO algorithm, are similarly approved in physical healthcare in IoT.

## 4.6 Conclusion

FDO, a modern meta-heuristic algorithm that simulates the reproduction behavior of the bee swarm in seeking better hives, was presented as one of the most recently formed algorithms. It was a very destructive algorithm when compared to other typical meta-heuristic algorithms since it performed exceptionally well throughout the optimization procedure. Rendering to the original results and outcomes in the FDO paper, the technique of the algorithm is quite powerful and outperforms other standard meta-heuristic algorithms.

An improvement in this chapter was finished to the FDO from two main viewpoints. Firstly; the case study for calculating FDO mathematical examples accurately explained each step as showing that the fitness global best solution for the second iteration, which is (**0**) is smaller than the first iteration, which is (**3188**). As a result, readers might just have a greater knowledge of the algorithm, which they could apply to solve real-world problems in the future. The second perspective is on an IoT-based healthcare application system that might be implemented in meta-heuristic algorithms to merely collect big-data behaviors. Moreover; it has been proposed that FDO parameters be adjusted to enhance the meta-heuristic evolutionary algorithm for evaluating large data in the IoT healthcare system. The future scope for this chapter recommends that FDO can be used for a virtualized health IoT architecture and highlighted the best big data system. This scope integrates the health cloud platform for telecommunications to engage and improve user quality and make the health IoT application more intimately related to general human beings.